\documentclass[runningheads]{llncs}
\usepackage[T1]{fontenc}
\usepackage{graphicx}
\usepackage{amsmath}
\usepackage{amssymb}
\usepackage{amsfonts}
\usepackage{tikz}
\usepackage{url}
\usepackage{hyperref}
\hypersetup{colorlinks=true,
            linkcolor=blue,
            anchorcolor=blue,
            citecolor=blue}
\usepackage{float}
\usepackage{subfig}
\usepackage{overpic}
\usepackage{booktabs}
\usepackage{multirow}
\usepackage[normalem]{ulem}
\useunder{\uline}{\ul}{}
\usepackage{bbding}

\usepackage{color}

\title{HRDecoder: High-Resolution Decoder Network for Fundus Image Lesion Segmentation}
\titlerunning{High-Resolution Decoder Network for Fundus Image Lesion Segmentation}

\author{Ziyuan~Ding\inst{1} 
\and Yixiong~Liang\inst{1}\Envelope\orcidID{0000-0003-0407-5838}  
\and Shichao~Kan\inst{1}
\and Qing~Liu\inst{2}
}
\authorrunning{Z. Ding et al.}
\institute{
Central South University, Hunan, China\\
\email{ \{ziyuanding, yxliang, kanshichao\}@csu.edu.cn}
\and 
University of Oulu, Oulu, Finland
}

\begin{document}

\maketitle              

\begin{abstract}

High resolution is crucial for precise segmentation in fundus images, yet handling high-resolution inputs incurs considerable GPU memory costs, with diminishing performance gains as overhead increases. To address this issue while tackling the challenge of segmenting tiny objects, recent studies have explored local-global fusion methods. These methods preserve fine details using local regions and capture long-range context information from downscaled global images. However, the necessity of multiple forward passes inevitably incurs significant computational overhead, adversely affecting inference speed. In this paper, we propose HRDecoder, a simple High-Resolution Decoder network for fundus lesion segmentation. It integrates a high-resolution representation learning module to capture fine-grained local features and a high-resolution fusion module to fuse multi-scale predictions. Our method effectively improves the overall segmentation accuracy of fundus lesions while consuming reasonable memory and computational overhead, and maintaining satisfying inference speed. Experimental results on the IDRiD and DDR datasets demonstrate the effectiveness of our method. Code is available at \url{https://github.com/CVIU-CSU/HRDecoder}.

\end{abstract}

\section{Introduction}
\begin{figure}[t]
    \centering
    \begin{minipage}[c]{0.6\textwidth}
        \centering
        \subfloat[
        Performance-cost Trade-off on DDR~\cite{DDR}\label{fig-1-a}
        ]{\includegraphics[width=\linewidth]{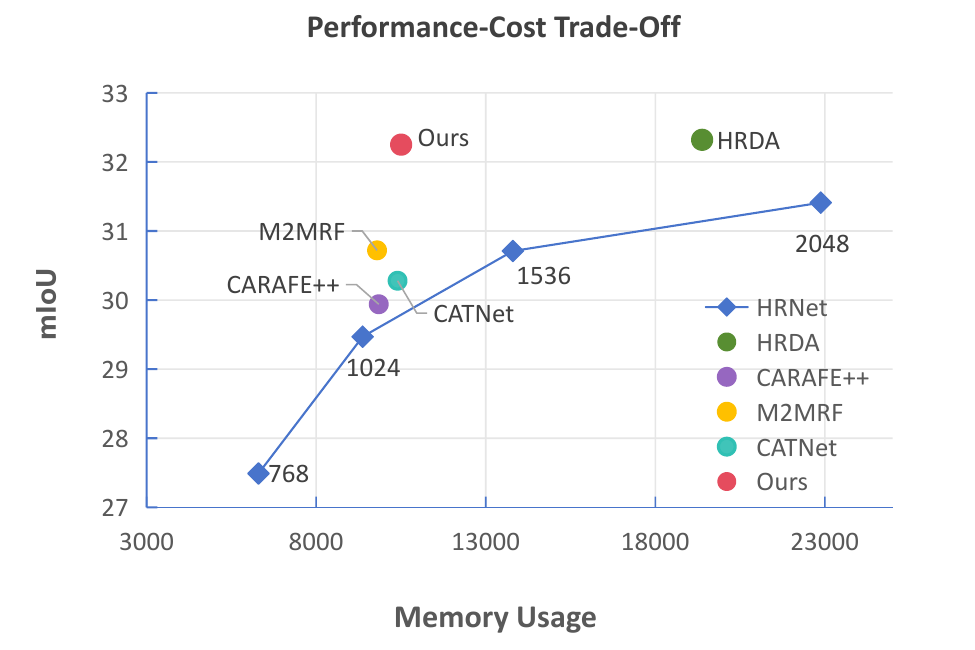}}
    \end{minipage}%
    \begin{minipage}{0.39\textwidth}
        \centering
        \subfloat[
        An example from IDRiD~\cite{IDRiD}\label{fig-1-b}
        ]{\includegraphics[width=\linewidth]{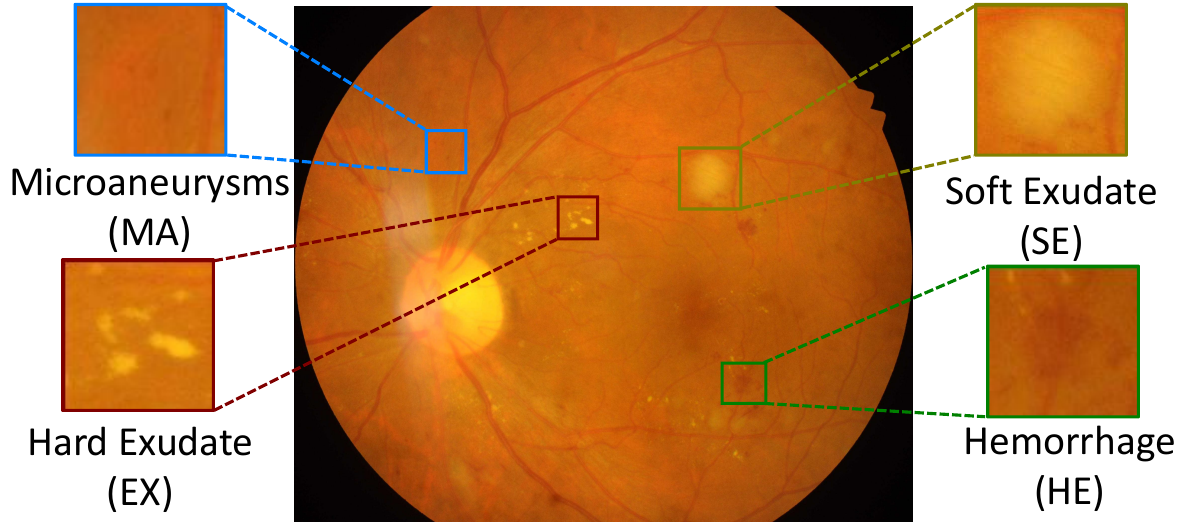}}\\
        \subfloat[
        Improvement on IDRiD~\cite{IDRiD}\label{fig-1-c}
        ]{\includegraphics[width=\linewidth]{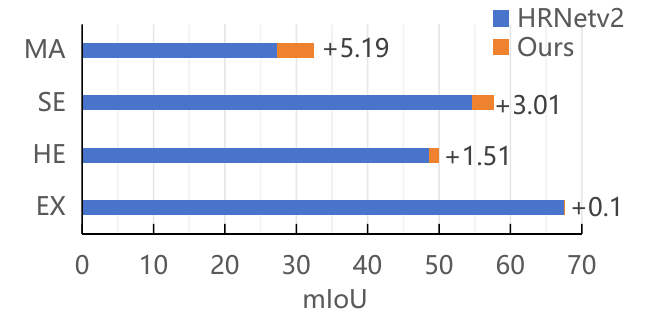}}
    \end{minipage}
    \caption{(a) Our method reaches SOTA performance and is memory efficient, the numbers represent input resolutions. (b) Example of tiny lesions in fundus images. (c) Performance gains on each category.}
    \label{fig-1}
\end{figure}

Fundus image lesion segmentation poses a significant challenge in medical image analysis, which is crucial for the early detection and monitoring of various retinal diseases.
The pixel-wise classification of tiny lesions, as shown in Fig.~\ref{fig-1-b}, demands considerably higher resolution compared to other segmentation tasks. 
Simply increasing the input resolution does boost segmentation performance of tiny lesions, while this is accompanied by rising memory usage, increasing computational overhead, and slower inference speed. 
These issues significantly hinder practical application and further performance improvement of models. 

Numerous efforts have been made to address the segmentation of tiny objects efficiently. 
Some studies explore FPN-like~\cite{FPN} or UNet-like~\cite{UNet} multi-scale features fusion methods to compensate for fine-grained details and enhance performance on small objects~\cite{APPNet,CATNet,CARAFE++,PMCNet,CPFNet,HiFormer,AFA}. 
However, these multi-scale feature fusion methods are susceptible to distortion from low-resolution features and result in suboptimal performance.
There are also other methods that emphasize local features~\cite{M2MRF,PCAA,SCS-Net,GLNet,HRDA}. 
M2MRF~\cite{M2MRF} enhances performance on tiny lesions by designing specialized local feature fusion modules, while it excessively emphasizes local features and exhibits slow convergence on larger lesions.
HRDA~\cite{HRDA} and GLNet~\cite{GLNet} design a dual-branch network composed of a High-Resolution(HR) branch to learn texture details from randomly cropped images and a Low-Resolution(LR) branch to extract contextual information through scaling operations, consequently achieving impressive results.
However, they need multiple forward passes, which significantly increase computational overhead and sharply decrease inference speed, thus restricting further applications.

Instead, we propose HRDecoder, a simple framework combining the idea of local-global high-resolution crops and multi-scale fusion at the decoder stage, to efficiently and effectively segment tiny lesions.
HRDecoder consists of an HR representation learning module to mine detailed features and an HR fusion module for integrating multi-scale predictions, significantly enhancing performance on small lesions (see Fig.~\ref{fig-1-c}).
By simply using scaling and cropping and a light-weight decoder, our method does not introduce extra parameters, and can significantly alleviate issues of high memory usage, computational overhead, and slow inference speed, as shown in Fig.~\ref{fig-1-a}.

We summarize the contributions as follows:
(1) We propose HRDecoder, a simple framework to address the challenge of segmenting tiny lesions in fundus images.
(2) Our method not only improves segmentation performance but also effectively mitigates high memory consumption, high computational overhead and slow inference speed.
(3) Our framework dose not introduce any extra trainable parameters and can be easily applied to existing methods.

\begin{figure}[t]
    \centering
    \includegraphics[width=1.0\textwidth]{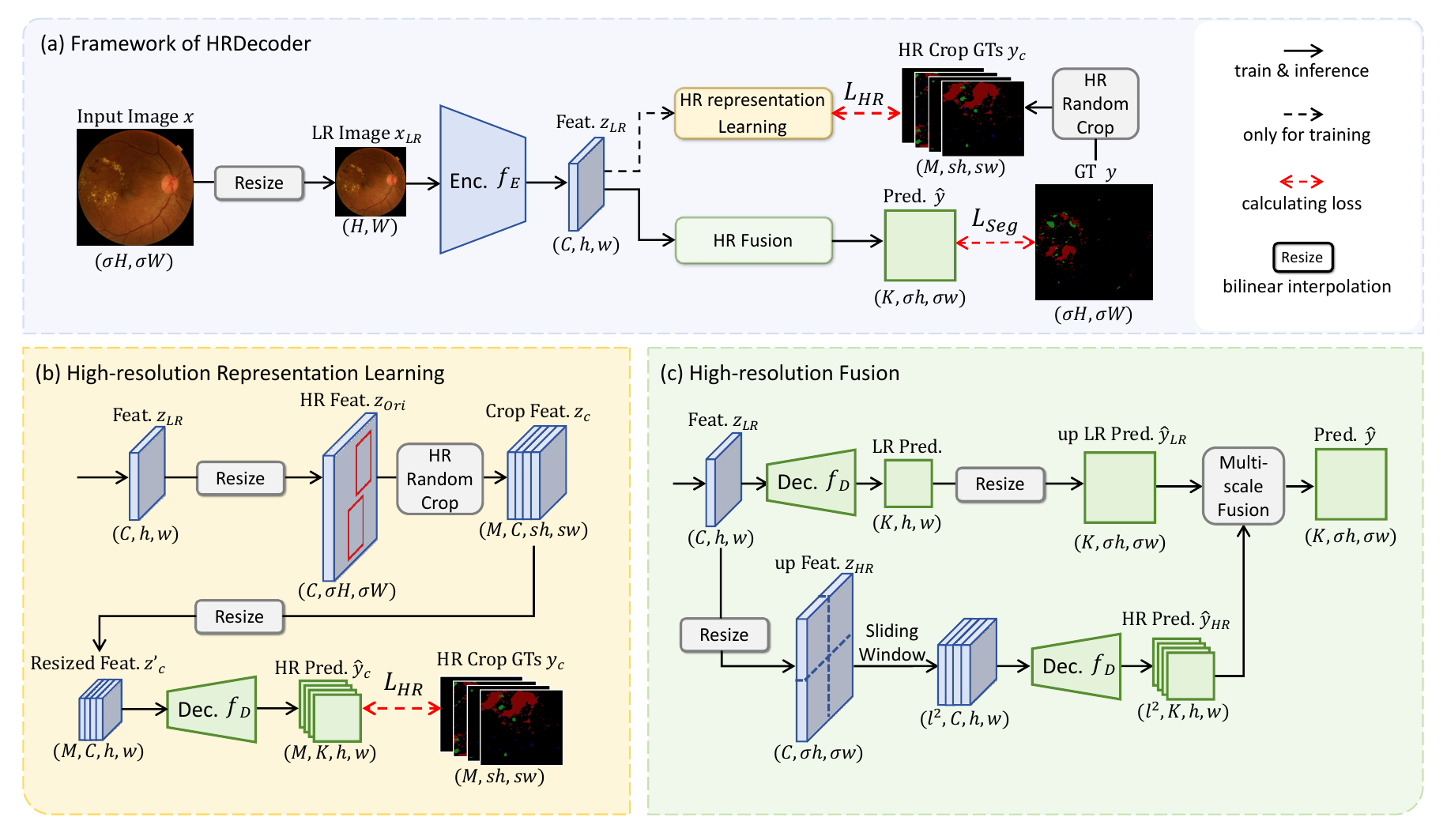}
    \caption{
    Overview of HRDecoder. (a) Training and testing pipeline. (b) HR representation learning module aims to learn local detailed features from simulated HR feature maps. (c) HR fusion module aggregates multi-scale predictions.}
    \label{fig-overview}
\end{figure}

\section{Method}

In this work, we propose HRDecoder for fundus image lesion segmentation. Our method consists of two modules: one that simulates HR inputs to enhance the representation learning of small objects (Sec.~\ref{sec-HRL}), and another that integrates HR multi-scale predictions to capture detailed information and local contextual cues for tiny targets (Sec.~\ref{sec-HFF}). Our approach can enhance performance at reasonable costs in terms of memory usage, overhead, and inference speed.

As shown in Fig.~\ref{fig-overview}a, given an input image 
$x\in\mathbb{R} ^ {3{\times}{\sigma}H{\times}{\sigma}W}$
with corresponding GT 
$y\in\mathbb{R} ^ {K\times \sigma H\times \sigma W}$ where $\sigma$$\geq$1 is the scale factor and $K$ is the number of classes, 
we first downsample $x$ to 
$x_{LR}\in \mathbb{R} ^ {3\times H\times W}$. 
LR features 
$z_{LR}$=$f_E(x_{LR})\in \mathbb{R} ^ {C\times h\times w}$ 
where $C$ is the number of channels are then extracted by the encoder $f_E$. 
Then, the HR representation learning module is adopted to enhance detail representation during training, and the HR fusion module for generating prediction 
$\hat{y}\in \mathbb{R} ^ {K\times \sigma h\times \sigma w}$ 
during training and inference. 

\subsection{High-resolution Representation Learning}
\label{sec-HRL}
In~\cite{HRDA}, HR local patches are randomly cropped out to maintain details and simultaneously the original image is resized into LR inputs to learn long-range context. The main drawback is that the image needs to go through the encoder $f_E$ multiple times, which incurs considerable memory consumption and computational overhead. Moreover, sliding window is used for preserving details during inference, resulting in a notable reduction in inference speed. Therefore, we introduce HR representation learning module to address above issues (see Fig.~\ref{fig-overview}b). 

We simply use a lightweight decoder to mine details from large-scale LR features.
Specifically, we first upsample feature $z_{LR}$ and obtain $z_{Ori}\in \mathbb{R} ^ {C\times \sigma H\times \sigma W}$, then random cropping is adopted to get $M$ patches of features $z_c$ and corresponding GT patches $y_c$. We follow~\cite{HRDA} to randomly sample $M$ bounding boxes $\{b_c^i\}_{i=0}^{M-1}$ from a discrete uniform distribution within the original size:
\begin{equation}\label{eq-1}
    \begin{aligned}
        z_c^i=z_{Ori}[b_{c,1}^i:b_{c,2}^i,b_{c,3}^i:b_{c,4}^i],
        \ \ y_c^i=y[b_{c,1}^i:b_{c,2}^i,b_{c,3}^i:b_{c,4}^i],
    \end{aligned}
\end{equation}
\begin{equation}\label{eq-2}
    \begin{aligned}
        {b}_{c,1}^i\sim \mathcal{U}\{ 0,\, (\sigma H-sh)/d\} \cdot d,\, \, \, \, {b}_{c,2}^i&={b}_{c,1}^i+sh, \\
        {b}_{c,3}^i\sim \mathcal{U}\{ 0,\, (\sigma W-sw)/d\} \cdot d,\, \, \, {b}_{c,4}^i&={b}_{c,3}^i+sw.
    \end{aligned}
\end{equation}
Here, we adjust the coordinates to be divided by $d$=8 in order to mitigate distortion of features and the crop ratio $s$ can be represented as (1-$\delta$,1+$\delta$) by the crop factor $\delta \in [0,1]$. We set $s$ to mimic random crop operation in preprocessing to enable the model to further learn lesion features at different scales. Later, $z_c$ will be scaled to a uniform size of $h\times w$ and obtain $z'_c$. With resized features $z'_c$, we calculate the HR loss $L_{HR}$ for cropped GT masks $y_c$, which is formulated as:
\begin{equation}\label{eq-3}
    \begin{aligned}
        L_{HR}&=\Sigma_{j=1}^{K}L_{Dice}(upsample(f_D(z'_c)^j),y_c^j),
    \end{aligned}
\end{equation}
where $L_{Dice}$ represents the binary Dice loss. With the supervision of $L_{HR}$, the model can progressively learn intricate local features from high-resolution GT, thereby mitigating distortions during upsampling and downsampling processes and enhancing segmentation performance, especially for tiny lesions. 

\subsection{High-resolution Fusion}
\label{sec-HFF}
The HR representation learning module empowers the extraction of intricate information from large-scale LR feature maps. Nevertheless, it tends to over emphasize localized tiny lesions and disregards larger ones. Thus, the small-scale LR feature map is essential to provide a holistic contextual understanding. To this end, we propose a simple HR fusion module to integrate multi-scale predictions to generate a more comprehensive and refined prediction.

Specifically, we follow~\cite{HRDA,GLNet} and adopt a dual-branch network. 
As shown in Fig.~\ref{fig-overview}c, the LR branch takes the LR feature $z_{LR}$ as input, feeds it into the decoder $f_D$ to obtain global prediction, which is then upsampled to $\hat{y}_{LR}$ by scale factor $\sigma$. 
This branch serves to provide rich global contextual information. 
Meanwhile, the HR branch takes in HR feature $z_{HR}$, also upscaled by $\sigma$, to capture detailed local information. 
We use a sliding window to get $l^2$ crops, where $l^2$ denotes the number of windows. 
Typically, we set both the stride and window size as ($h,w$). 
HR prediction $\hat{y}_{HR}$ is later obtained by aggregating the results from sliding windows. 
With $\hat{y}_{LR}$ and $\hat{y}_{HR}$, we fuse them together to generate the final prediction $\hat{y}$:
\begin{equation}\label{eq-4}
    \begin{aligned}
        \hat{y} = f_A(z)\odot\hat{y}_{HR} + (1-f_A(z))\odot\hat{y}_{LR}.
    \end{aligned}
\end{equation}
Here, we use $f_A(\cdot)$ to denote a general form of scale attention, and $z$ is used to denote the features of the branch from which the scale attention is to be learned. In HRDA~\cite{HRDA}, it represents using an additional decoder head $f_A$ to learn a scale attention on the LR branch. While in our method, we simply design $f_A$ as a fixed value of 0.5. Given that most regions in fundus images are background areas, learning scale attention from LR branch may hinder the ability to capture a coherent scene layout, while weighted sum can better grasp lesion details and contextual information (further discussed in Tab.~\ref{tab-3}).
With the fused prediction $\hat{y}$, we can calculate the segmentation loss: 
\begin{equation}
    \begin{aligned}
        L_{Seg}&=\Sigma_{j=1}^{K}L_{Dice}(upsample(\hat{y}^j),y^j).
    \end{aligned}
\end{equation}
Eventually, the overall loss function is formulated as:
\begin{equation}
    \begin{aligned}
        L&=L_{Seg}+\lambda L_{HR},
    \end{aligned}
\end{equation}
where $\lambda$ is hyper-parameter and is empirically set to 0.1.

By employing these two modules, HRDecoder learns detailed features and retains contextual information using small-scale inputs through the encoder and large-scale feature maps through the decoder.
By simply using interpolation and cropping operations, our method does not introduce extra parameters and can be easily applied to existing methods. 
Furthermore, given that the resource overhead of the decoder is significantly lower than that of the encoder, HRDecoder can effectively alleviate issues such as high memory usage, computational overhead, and slow inference speeds.

\section{Experiment}
\subsection{Datasets and Implementation Details}
\subsubsection{Datasets:}
We conduct experiments on two main public retinal lesion segmentation datasets, i.e. IDRiD~\cite{IDRiD} and DDR~\cite{DDR}. The IDRiD dataset contains 81 high-quality retinal lesion segmentation images with a unified resolution of 4288$\times$2848, 54 for training and 27 for testing. The DDR dataset consists of 757 color fundus images, with 383 for training, 149 for validation and the rest 225 for testing. The image resolutions vary from 1088$\times$1920 to 3456$\times$5184. Both datasets provide pixel-level annotations for four different lesions, i.e. hard exudates (EX), hemorrhages (HE), soft exudates (SE), and microaneurysms (MA).

\subsubsection{Implementation Details:}
HRDecoder is implemented based on the MMSegmentation~\cite{MMSeg} framework. We adopt HRNetv2~\cite{HRNet} as backbone and simple FCNHead~\cite{FCN} as decoder. Images from IDRiD and DDR datasets are resized to 1440$\times$960 and 1024$\times$1024 in previous protocols, and we set scale factor $\sigma$ to 2, i.e. 2880$\times$1920 for IDRiD and 2048$\times$2048 for DDR, respectively. To reduce memory cost during training on IDRiD dataset, we randomly crop images to 1920$\times$1920 and sliding window is used for inference. SGD with a learning rate of 0.01 is used for optimization. Total batch size is 4, and iterations are set to 20k for IDRiD and 40k for DDR. For hyper-parameters, the crop number $M$ is set to 2 and 4, respectively. We set the crop factor $\delta$ to 0.25 to implicitly learn multi-scale features. For evaluation metrics, we follow~\cite{M2MRF,Bi-VLGM} and utilize the commonly used IoU, F-score, AUPR and their mean values.

\begin{table}[t]
\centering
\caption{Comparison with previous SOTA methods on IDRiD~\cite{IDRiD} and DDR~\cite{DDR} datasets. $^\dag$\ represents we reproduce it with the same experimental settings as ours. * means we implement our method with ConvNeXt~\cite{ConvNeXt} backbone. Results are averaged over three repetitions. }\label{tab-1}
\resizebox{1.0\linewidth}{!}{
    \begin{tabular}{c|ccccc|ccccc|ccccc}
    \toprule
    \multirow{2}{*}{Methods} & \multicolumn{5}{c|}{AUPR} & \multicolumn{5}{c|}{F} & \multicolumn{5}{c}{IoU} \\ \cline{2-16} 
     & EX & HE & SE & \multicolumn{1}{c|}{MA} & mAUPR & EX & HE & SE & \multicolumn{1}{c|}{MA} & mF & EX & HE & SE & \multicolumn{1}{c|}{MA} & mIoU \\ 
    \bottomrule
    \multicolumn{16}{c}{IDRiD} \\
    \toprule
    DNL~\cite{DNL} & 75.12 & 64.04 & 64.73 & \multicolumn{1}{c|}{32.48} & 59.09 & 73.15 & 61.87 & 63.96 & \multicolumn{1}{c|}{32.78} & 57.94 & 57.67 & 44.80 & 47.03 & \multicolumn{1}{c|}{19.61} & 42.28 \\
    HRNetv2$^\dag$~\cite{HRNet} & 82.75 & 67.85 & 71.96 & \multicolumn{1}{c|}{44.01} & 66.64 & {\ul 80.56} & 65.35 & 70.36 & \multicolumn{1}{c|}{42.97} & 64.81 & {\ul 67.44} & 48.53 & 54.62 & \multicolumn{1}{c|}{27.36} & 49.49 \\
    Swin$^\dag$~\cite{Swin} & {\ul 85.39} & 68.67 & 74.28 & \multicolumn{1}{c|}{43.47} & 67.95 & 77.57 & 65.36 & 70.88 & \multicolumn{1}{c|}{45.59} & 64.85 & 63.32 & 48.22 & 54.77 & \multicolumn{1}{c|}{29.58} & 48.97 \\
    Segformer$^\dag$~\cite{SegFormer} & 82.47 & 69.23 & 73.69 & \multicolumn{1}{c|}{37.63} & 65.75 & 76.10 & 64.55 & 69.49 & \multicolumn{1}{c|}{40.65} & 62.70 & 61.43 & 47.66 & 53.24 & \multicolumn{1}{c|}{25.52} & 46.96 \\
    Mask2Former$^\dag$~\cite{Mask2Former} & 84.68 & 69.01 & 74.88 & \multicolumn{1}{c|}{42.53} & 67.78 & 79.64 & 66.47 & 68.05 & \multicolumn{1}{c|}{44.64} & 64.70 & 66.17 & 49.78 & 51.59 & \multicolumn{1}{c|}{28.75} & 49.07 \\
    IFA~\cite{IFA} & 81.92 & 69.01 & 70.47 & \multicolumn{1}{c|}{46.35} & 66.94 & 79.80 & \textbf{67.43} & 69.12 & \multicolumn{1}{c|}{46.35} & 65.68 & 66.39 & \textbf{50.86} & 52.82 & \multicolumn{1}{c|}{30.17} & 50.06 \\
    PCAA~\cite{PCAA} & 81.63 & 66.74 & 75.49 & \multicolumn{1}{c|}{43.33} & 66.80 & 79.58 & 64.59 & \textbf{74.13} & \multicolumn{1}{c|}{43.17} & 65.37 & 66.09 & 47.70 & \textbf{58.89} & \multicolumn{1}{c|}{27.53} & 50.05 \\
    TGANet~\cite{TGANet} & 82.16 & 65.60 & 68.86 & \multicolumn{1}{c|}{42.19} & 64.70 & 80.01 & 63.46 & 67.89 & \multicolumn{1}{c|}{41.29} & 63.16 & 66.67 & 46.48 & 51.39 & \multicolumn{1}{c|}{26.01} & 47.64 \\
    LViT~\cite{LVit} & 82.19 & 63.36 & 70.32 & \multicolumn{1}{c|}{43.65} & 64.88 & 79.99 & 60.96 & 69.33 & \multicolumn{1}{c|}{43.44} & 63.43 & 66.65 & 43.85 & 53.06 & \multicolumn{1}{c|}{27.74} & 47.82 \\
    M2MRF$^\dag$~\cite{M2MRF} & 82.10 & 67.96 & 71.77 & \multicolumn{1}{c|}{{\ul 46.83}} & 67.17 & 79.81 & 65.93 & 70.36 & \multicolumn{1}{c|}{46.28} & 65.60 & 66.40 & 49.18 & 54.31 & \multicolumn{1}{c|}{30.18} & 50.02 \\
    ConvNeXt$^\dag$~\cite{ConvNeXt} & 83.96 & \textbf{72.64} & {\ul 77.12} & \multicolumn{1}{c|}{45.97} & {\ul 69.93} & 76.52 & 68.17 & 72.82 & \multicolumn{1}{c|}{{\ul 47.48}} & 66.25 & 61.98 & 51.72 & 57.26 & \multicolumn{1}{c|}{{\ul 31.13}} & 50.52 \\
    Bi-VLGM~\cite{Bi-VLGM} & 82.48 & 69.32 & 74.50 & \multicolumn{1}{c|}{46.20} & 68.13 & 80.51 & {\ul 67.42} & 72.95 & \multicolumn{1}{c|}{45.98} & {\ul 66.71} & 67.38 & {\ul 50.85} & 57.41 & \multicolumn{1}{c|}{29.85} & {\ul 51.37} \\ \hline
    HRDecoder & \textbf{87.55} & {\ul 70.80} & \textbf{77.65} & \multicolumn{1}{c|}{\textbf{49.16}} & \textbf{71.29} & \textbf{80.61} & 66.68 & {\ul 72.99} & \multicolumn{1}{c|}{\textbf{49.35}} & \textbf{67.41} & \textbf{67.54} & 50.04 & {\ul 57.63} & \multicolumn{1}{c|}{\textbf{32.55}} & \textbf{51.94} \\
    HRDecoder* & 84.65 & 71.53 & 76.23 & \multicolumn{1}{c|}{50.28} & 70.67 & 79.34 & 68.45 & 72.89 & \multicolumn{1}{c|}{50.58} & 67.82 & 66.10 & 52.26 & 57.45 & \multicolumn{1}{c|}{33.96} & 52.44 \\ 
    \bottomrule
    \multicolumn{16}{c}{DDR} \\
    \toprule
    DNL~\cite{DNL} & 56.05 & 47.81 & 42.01 & \multicolumn{1}{c|}{14.71} & 40.15 & 53.36 & 42.71 & 40.40 & \multicolumn{1}{c|}{15.60} & 38.02 & 36.39 & 27.15 & 25.33 & \multicolumn{1}{c|}{8.46} & 24.33 \\
    HRNetv2$^\dag$~\cite{HRNet} & 61.48 & 51.01 & 47.42 & \multicolumn{1}{c|}{24.62} & 46.13 & 58.78 & 48.95 & 46.17 & \multicolumn{1}{c|}{24.61} & 44.63 & 41.58 & 32.32 & 29.98 & \multicolumn{1}{c|}{13.98} & 29.47 \\
    Swin$^\dag$~\cite{Swin} & {\ul 64.49} & 55.80 & 50.07 & \multicolumn{1}{c|}{19.51} & 47.47 & 60.05 & 53.46 & 47.96 & \multicolumn{1}{c|}{25.97} & 46.86 & 42.91 & 36.50 & 31.23 & \multicolumn{1}{c|}{14.92} & 31.39 \\
    Segformer$^\dag$~\cite{SegFormer} & 61.43 & 52.53 & 33.88 & \multicolumn{1}{c|}{19.88} & 41.93 & 57.49 & 46.47 & 32.16 & \multicolumn{1}{c|}{24.73} & 40.21 & 40.34 & 30.30 & 19.18 & \multicolumn{1}{c|}{14.11} & 25.98 \\
    Mask2Former$^\dag$~\cite{Mask2Former} & 63.28 & 55.26 & {\ul 51.87} & \multicolumn{1}{c|}{19.03} & 47.36 & 59.60 & 49.73 & \textbf{53.36} & \multicolumn{1}{c|}{23.63} & 46.58 & 42.45 & 33.35 & \textbf{36.28} & \multicolumn{1}{c|}{13.40} & 31.37 \\
    IFA~\cite{IFA} & 61.51 & 46.19 & 48.90 & \multicolumn{1}{c|}{12.98} & 42.40 & 56.76 & 46.28 & 48.25 & \multicolumn{1}{c|}{0.55} & 37.96 & 39.62 & 30.11 & 31.80 & \multicolumn{1}{c|}{0.28} & 25.45 \\
    PCAA~\cite{PCAA} & 60.57 & 57.46 & 41.49 & \multicolumn{1}{c|}{18.58} & 44.53 & 56.89 & {\ul 54.47} & 36.68 & \multicolumn{1}{c|}{20.57} & 42.15 & 39.76 & {\ul 37.43} & 22.46 & \multicolumn{1}{c|}{11.47} & 27.78 \\
    TGANet~\cite{TGANet} & 60.49 & 52.63 & 43.55 & \multicolumn{1}{c|}{26.81} & 45.87 & 58.92 & 42.19 & 41.27 & \multicolumn{1}{c|}{26.92} & 42.33 & 41.76 & 26.73 & 26.00 & \multicolumn{1}{c|}{15.55} & 27.51 \\
    LViT~\cite{LVit} & 61.35 & 46.29 & 48.06 & \multicolumn{1}{c|}{{\ul 27.61}} & 45.83 & 59.15 & 42.85 & 46.88 & \multicolumn{1}{c|}{{\ul 27.78}} & 44.17 & 42.00 & 27.27 & 30.62 & \multicolumn{1}{c|}{{\ul 16.13}} & 29.01 \\
    M2MRF$^\dag$~\cite{M2MRF} & 63.74 & 54.88 & 49.95 & \multicolumn{1}{c|}{\textbf{27.91}} & 49.12 & {\ul 60.41} & 47.60 & 48.73 & \multicolumn{1}{c|}{27.70} & 46.11 & {\ul 43.28} & 31.27 & 32.25 & \multicolumn{1}{c|}{16.08} & \multicolumn{1}{l}{30.72} \\
    ConvNeXt$^\dag$~\cite{ConvNeXt} & 63.87 & \textbf{58.03} & 50.61 & \multicolumn{1}{c|}{15.89} & 47.10 & 58.71 & \textbf{55.18} & 50.91 & \multicolumn{1}{c|}{22.22} & 46.76 & 41.50 & \textbf{38.05} & 34.08 & \multicolumn{1}{c|}{12.37} & \multicolumn{1}{l}{31.50} \\
    Bi-VLGM~\cite{Bi-VLGM} & 62.01 & {\ul 57.38} & 50.95 & \multicolumn{1}{c|}{26.19} & {\ul 49.13} & 57.90 & 54.38 & 50.81 & \multicolumn{1}{c|}{26.06} & {\ul 47.29} & 40.75 & 37.34 & 34.06 & \multicolumn{1}{c|}{14.98} & \multicolumn{1}{l}{{\ul 31.78}} \\ \hline
    HRDecoder & \textbf{64.84} & 55.69 & \textbf{51.93} & \multicolumn{1}{c|}{24.60} & \textbf{49.27} & \textbf{60.67} & 52.70 & {\ul 51.57} & \multicolumn{1}{c|}{\textbf{27.92}} & \textbf{48.21} & \textbf{43.54} & 35.03 & {\ul 34.22} & \multicolumn{1}{c|}{\textbf{16.22}} & \multicolumn{1}{l}{\textbf{32.25}} \\
    HRDecoder* & 63.73 & 55.40 & 52.70 & \multicolumn{1}{c|}{21.08} & 48.20 & 59.81 & 53.67 & 53.56 & \multicolumn{1}{c|}{26.44} & 48.37 & 42.67 & 36.67 & 36.57 & \multicolumn{1}{c|}{15.17} & \multicolumn{1}{l}{32.77} \\ 
    \bottomrule
    \end{tabular}
}
\end{table}

\subsection{Comparison with State-of-the-Art (SOTA) Methods}
\subsubsection{SOTA Segmentation Methods:}
First, we present a comprehensive comparison between our method and previous SOTA approaches including different backbones~\cite{HRNet,Swin,SegFormer,ConvNeXt}, feature-enhanced methods~\cite{DNL,Mask2Former,PCAA,M2MRF} and multi-modal methods~\cite{IFA,TGANet,LVit,Bi-VLGM} on IDRiD~\cite{IDRiD} and DDR~\cite{DDR} datasets in Tab.~\ref{tab-1}. The best and second-best scores are marked in bold and underlined, respectively. We primarily report results utilizing HRNet as backbone for fair comparison. For IDRiD dataset, HRDecoder achieves results of 71.29\%, 67.41\%, and 51.94\% in mAUPR, mF, and mIoU, respectively. Our method outperforms previous SOTA method~\cite{Bi-VLGM} by a large margin of 3\% in mAUPR. In terms of each lesion, our method achieves the best or second-best in 3 out of 4 in AUPR, F and IoU. For DDR dataset, consistent conclusions can be drawn as on IDRiD. 

We observe that ConvNeXt~\cite{ConvNeXt} and Swin~\cite{Swin,Mask2Former} exhibit relatively good performance. We assert that CNN models or local window attention can effectively capture intricate features within localized regions. In contrast, attention-based methods~\cite{SegFormer,SenFormer} struggle to extract detailed information from fundus images due to long-range dependency mechanism. Therefore, a simple CNN decoder is helpful in capturing fine-grained features efficiently and effectively. Benefiting from the high-resolution design in~\cite{HRNet} and our exploration of HR feature maps, our method significantly improves performance on tiny lesions within fundus images. Visual results on IDRiD and DDR are provided in supplementary material.

\begin{table}[t]
\centering
\caption{
Comparison with different multi-scale fusion methods on IDRiD~\cite{IDRiD} and DDR~\cite{DDR} testing set. $^\ddag$\ means the result is from~\cite{PMCNet} or~\cite{Bi-VLGM}. Results are averaged over three repetitions.
}\label{tab-2}
\resizebox{1.0\linewidth}{!}{
    \begin{tabular}{c|ccc|ccc|c|c|c|c}
    \toprule
    \multirow{2}{*}{Methods} & \multicolumn{3}{c|}{IDRiD} & \multicolumn{3}{c|}{DDR} & \multirow{2}{*}{Params(M)} & \multirow{2}{*}{GFLOPs} & \multirow{2}{*}{Memory(GB)} & \multirow{2}{*}{FPS(img/s)} \\ \cline{2-7}
     & mAUPR & mF & mIoU & mAUPR & mF & mIoU &  &  &  &  \\ \hline
    HRNet-1024~\cite{HRNet} & 66.64 & 64.81 & 49.49 & 46.13 & 44.63 & 29.57 & 65.57 & 355.62 & 9.2 & 6.92 \\
    HRNet-2048~\cite{HRNet} & 69.37 & 66.87 & 51.45 & 48.56 & 46.96 & 31.41 & 65.57 & 1422.48 & 22.3 & 2.36 \\ \hline
    CARAFE++~\cite{CARAFE++} & 67.56 & 65.52 & 49.92 & 47.48 & 45.27 & 29.94 & 66.13 & 481.71 & 9.6 & 4.88 \\
    PMCNet$^\ddag$~\cite{PMCNet} & 68.08 & 56.02 & 43.12 & 36.44 & 39.31 & 32.29 & -- & -- & -- & -- \\
    SenFormer~\cite{SenFormer} & 66.36 & 63.04 & 47.35 & 43.80 & 41.97 & 27.36 & 68.81 & 413.43 & 9.7 & 5.45 \\
    CATNet~\cite{CATNet} & 67.36 & 65.82 & 50.34 & 45.74 & 45.57 & 30.28 & 67.34 & 383.05 & 10.2 & 7.23 \\ \hline
    PCAA$^\ddag$~\cite{PCAA} & 66.80 & 65.37 & 50.05 & 44.52 & 52.15 & 27.78 & -- & -- & -- & -- \\
    M2MRF~\cite{M2MRF} & 67.17 & 65.60 & 50.02 & 49.12 & 46.11 & 30.72 & 70.30 & 353.65 & 9.6 & 7.56 \\
    HRDA~\cite{HRDA} & 71.17 & 67.45 & 52.09 & 49.13 & 47.95 & 32.32 & 66.06 & 2113.45 & 18.9 & 1.52 \\ 
    HRDecoder & 71.29 & 67.41 & 51.94 & 49.27 & 47.81 & 32.25 & 65.57 & 420.42 & 10.3 & 4.86 \\
    \bottomrule
    \end{tabular}
}
\end{table}

\subsubsection{Multi-scale Fusion Methods:}
To demonstrate the effectiveness and efficiency of HRDecoder, we compare various multi-scale fusion methods in terms of performance, parameter, computational overhead, GPU memory, and inference speed in Tab.~\ref{tab-2}. We report results of HRNet~\cite{HRNet} at different scales in the first group. The second group~\cite{CARAFE++,PMCNet,SenFormer,CATNet} employs FPN-like~\cite{FPN} global feature fusion methods. The third group~\cite{PCAA,M2MRF,HRDA} utilizes local fusion methods.

Global feature fusion methods integrate features across multiple scales and improve segmentation performance. 
However, they are generally less effective than local fusion methods. The HRDA~\cite{HRDA} can significantly improve segmentation performance. 
However, it requires 5$\times$GFLOPs and 2$\times$GPU memory compared with above methods due to multiple forwards of encoder. 
Additionally, the sliding window notably reduces its inference speed. 
In contrast, our method can achieve comparable performance to HRDA without significantly increasing computational overhead or memory usage. Overall, HRDecoder can achieve a better performance-cost trade-off.

\begin{figure}[ht]
    \centering
    \subfloat[\label{fig-3-a}]{\includegraphics[width=0.34\textwidth]{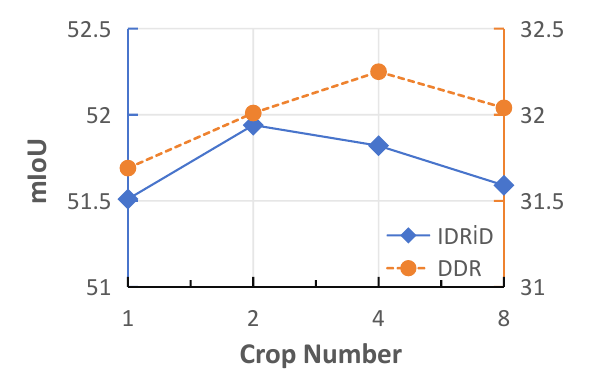}}
    \hfil
    \subfloat[\label{fig-3-b}]{\includegraphics[width=0.32\textwidth]{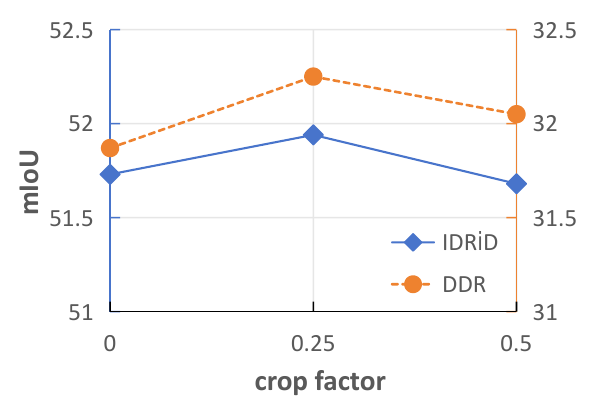}}
    \hfil
    \subfloat[\label{fig-3-c}]{\includegraphics[width=0.33\textwidth]{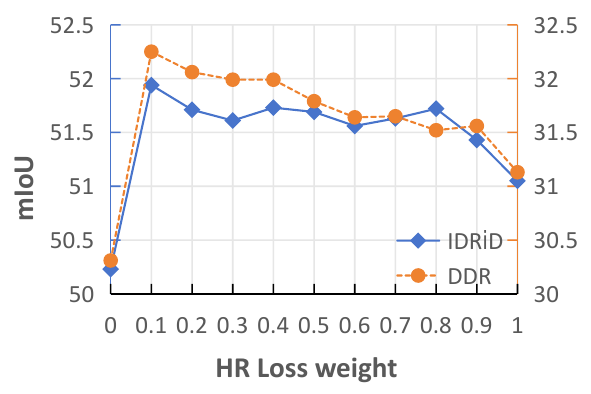}}
    \caption{Ablation on HR representation learning module. (a) Influence of the number of HR crops $M$. (b) Impact of crop factor $\delta$. (c) Impact of HR loss weight $\lambda$.}
    \label{fig-3}
\end{figure}
\noindent
\begin{minipage}[t]{0.45\textwidth}
\centering
\setlength{\abovecaptionskip}{0cm}
\captionof{table}{
Comparison of different fusion method on IDRiD test set.
}\label{tab-3}
\resizebox{1.0\linewidth}{!}{
    \begin{tabular}{c|c|c|c}
    \toprule
    Fusion Method $f_A(\cdot)$& mAUPR & mF & mIoU \\ \hline
    LR Attn. & 70.34 & 67.06 & 51.60 \\
    HR Attn. & 71.02 & 67.35 & 51.83 \\
    Weighted Sum & 71.29 & 67.41 & 51.94\\
    \bottomrule
    \end{tabular}
}
\end{minipage}
\hfil
\begin{minipage}[t]{0.45\textwidth}
\setlength{\abovecaptionskip}{0cm}
\centering
\captionof{table}{
Influence of HR scale factor $\sigma$ on IDRiD test set.
}\label{tab-4}
\resizebox{1.0\linewidth}{!}{
    \begin{tabular}{c|cccc|c}
    \toprule
    Scale Factor $\sigma$ & EX & HE & SE & MA & mIoU \\ \hline
    1$\times$ & 67.44 & 48.53 & 54.62 & 27.36 & 49.49 \\
    2$\times$ & 67.54 & 50.04 & 57.63 & 32.55 & 51.94 \\
    4$\times$ & 70.06 & 45.63 & 0.00 & 34.49 & 37.55 \\
    \bottomrule
    \end{tabular}
}
\end{minipage}

\subsection{Ablation Study}
Fig.~\ref{fig-3} shows the ablation study on HR representation learning module. Fig.~\ref{fig-3-a} shows the impact of the number of HR crops. Training with larger $M$ enhances the ability to capture details, while too many HR crops may lead the model to focus excessively on tiny lesions e.g. MA and neglect larger lesions e.g. SE. Hence, we set $M$ to 2 and 4 for IDRiD and DDR, respectively.
Fig.~\ref{fig-3-b} illustrates the influence of crop ratio (1-$\delta$,1+$\delta$). A larger $\delta$ results in a larger ratio range of crop size. Implicitly inputting multi-scale features enables the model learn features of various scales to address lesions of different sizes. We set $\delta$=0.25, as overly large $\delta$ may introduce excessive uncertainty.
Fig.~\ref{fig-3-c} shows the sensitivity to HR loss weight $\lambda$. A larger $\lambda$ tends to emphasize local features and diminish contextual information. Experimental results show that 0.1 is a preferable choice.

We also conduct ablation studies on HR fusion module. In Tab.~\ref{tab-3} we compare three different fusion strategies: learning scale attention from LR branch, from HR branch and weighted fusion. 
In HRDA~\cite{HRDA}, attention is learned from LR branch using an attention head, which is heavily affected by the predominance of background regions in fundus images. 
Instead, learning attention from HR detailed features or simple weighted fusion is favored and we adopt the simple parameter-free method.
Tab.~\ref{tab-4} demonstrates the effect of scale factor $\sigma$. $\sigma$=1 corresponds to the vanilla HRNet~\cite{HRNet}. When $\sigma$ is set too large, HR crops will excessively focus on tiny lesions e.g. EX and MA. However, for slightly larger lesions e.g. SE, the model may misclassify due to severe lack of local context.
Further ablation studies are provided in supplementary material.

\section{Conclusion}

In this paper, we propose HRDecoder, a straightforward framework that combines an HR representation learning module to mine local features and an HR fusion module to capture contextual information, for fundus image lesion segmentation. Our approach strikes a balance between performance and memory usage, inference speed, and overhead without introducing extra parameters. Overall, HRDecoder can achieve SOTA performance while maintaining manageable overhead. There are certain limitations: the simple feature fusion is specifically designed for fundus images, whereas scale attention may have broader applicability in various tasks. Additionally, the simple CNN-based decoder may fall short in capturing contextual information. Despite these limitations, we believe that our approach provides a simple and versatile solution.

\begin{credits}
\subsubsection{\ackname} This manuscript was supported in part by the National Key Research and Development Program of China under Grant 2021YFF1201202, the Natural Science Foundation of Hunan Province under Grant 2024JJ5444 and 2023JJ30699, the Key Research and Development Program of Hunan Province under Grant 2023SK2029, the Changsha Municipal Natural Science Foundation under Grant kq2402227, and the Research Council of Finland (formerly Academy of Finland for Academy Research Fellow project) under Grant 355095. The authors wish to acknowledge High Performance Computing Center of Central South University for computational resources.

\subsubsection{\discintname}
The authors have no competing interests to declare that are relevant to the content of this article.
\end{credits}

\bibliographystyle{splncs04}
\bibliography{Paper-4020}

\end{document}